\documentclass{article}
\usepackage{arxiv}

\usepackage[utf8]{inputenc} 
\usepackage[T1]{fontenc}    
\usepackage{hyperref}       
\usepackage{url}            
\usepackage{booktabs}       
\usepackage{amsfonts}       
\usepackage{nicefrac}       
\usepackage{microtype}      
\usepackage{lipsum}		
\usepackage{graphicx}
\usepackage[numbers]{natbib}
\usepackage{doi}
\usepackage{xcolor}
\usepackage{multirow}
\usepackage{float}

\title{A novel language model for predicting serious adverse event results in clinical trials from their prospective registrations}

\author
{Qixuan Hu~$^{1}$, Xumou Zhang~$^{1}$, Jinman Kim~$^{1}$, Florence Bourgeois~$^{2, 3}$, Adam G.~Dunn~$^{4}$\\ \\
\normalfont{\small $^{1}$School of Computer Science, Faculty of Engineering, University of Sydney, Sydney, NSW, Australia}\\
\normalfont{\small $^{2}$Computational Health Informatics Program, Boston Children’s Hospital, Boston, MA, United States}\\
\normalfont{\small $^{3}$Harvard-MIT Center for Regulatory Science and Department of Pediatrics, Harvard Medical School, Boston, MA, United States}\\
\normalfont{\small $^{4}$Sydney School of Public Health, Faculty of Medicine and Health, University of Sydney, Sydney, NSW, Australia} \vspace{2em}
}

\date{}

\renewcommand{\shorttitle}{\textit{arXiv} Template}

\hypersetup{
pdftitle={Cody Manuscript},
pdfauthor={Qixuan Hu, Xumou Zhang, Jinman Kim, Florence Bourgeois, Adam G.~Dunn},
pdfkeywords={Language model, clinical trial as a research topic, adverse events},
}

\begin{document}
\pagestyle{plain}  

\maketitle

\begin{abstract}
\textbf{Objectives}: With accurate estimates of expected safety results, clinical trials could be better designed and monitored. We evaluated methods for predicting serious adverse event (SAE) results in clinical trials using information only from their registrations prior to the trial.

\textbf{Material and Methods}: We analyzed 22,107 two-arm parallel interventional clinical trials from ClinicalTrials.gov with structured summary results. Two prediction models were developed: a classifier predicting whether a greater proportion of participants in an experimental arm would have SAEs (area under the receiver operating characteristic curve; AUC) compared to the control arm, and a regression model to predict the proportion of participants with SAEs in the control arms (root mean squared error; RMSE). A transfer learning approach using pretrained language models (e.g., ClinicalT5, BioBERT) was used for feature extraction, combined with a downstream model for prediction. To maintain semantic representation in long trial texts exceeding localized language model input limits, a sliding window method was developed for embedding extraction.

\textbf{Results}: The best model (ClinicalT5+Transformer+MLP) had 77.6\% AUC when predicting which trial arm had a higher proportion of SAEs. When predicting SAE proportion in the control arm, the same model achieved RMSE of 18.6\%. The sliding window approach consistently outperformed direct comparisons. Across 12 classifiers, the average absolute AUC increase was 2.00\%, and absolute RMSE reduction was 1.58\% across 12 regressors.

\textbf{Discussion}: Summary results data from ClinicalTrials.gov remains underutilized. Predicted results of publicly reported trials provides an opportunity to identify discrepancies between expected and reported safety results.

\end{abstract}


\section{Background}
\label{sec:background}
Clinical trials are designed to establish evidence of the efficacy and safety of new treatments and track their effectiveness and monitor their safety post-approval\cite{who2018registries}. Clinical trials are required to monitor participants for adverse events and report these results\cite{fda2023sae}, but data about adverse events are underutilized even where summary results are made publicly available. 

Clinical trials are increasingly required to be registered prospectively—before enrolling participants\cite{icmje_clinical_trials}. This is because trial registrations have improved transparency in reporting, the ability to identify reporting biases, and evidence synthesis\cite{tse2009reporting, chan2025reporting, lindsley2022clinical}. Clinical trial registrations include details of their purpose, who will be recruited, the treatments that are being tested, and what is being measured. ClinicalTrials.gov is the largest individual registry, and it includes clinical trial registration information for hundreds of thousands of trials\cite{mccray2000better}. ClinicalTrials.gov was updated in 2008 to include semi-structured summary results\cite{zarin2011clinicaltrials}, with 21\% of interventional studies (61,330 of 292,133) having reported these by March 2024. These summary results include participant flow, demographic and baseline characteristics, primary and secondary outcome measures, and adverse events\cite{hhs2017finalrule}. Adverse event reporting in ClinicalTrials.gov adheres to a consistent format, and this includes the number of participants experiencing SAEs in each study arm.

Taking advantage of the semi-structured nature of clinical trial registration data, there has been broad application of machine learning on the data. Prior work using ClinicalTrials.gov as the basis for training machine learning models to predict trial termination\cite{elkin2021predictive}, publication likelihood\cite{wang2022predicting}, new therapeutics progressing through trial phases towards approval\cite{feijoo2020key}, and trial durations for specific diseases\cite{long2023predicting}.  Despite the importance of clinical trials and their registrations for establishing evidence of safety, we know of no studies that have sought to predict safety related results of trials from registrations, such as serious adverse events. 

Predictive models of SAE results for clinical trials may be of value for a range of novel use cases. For example, by predicting the expected SAE results of a trial, it may be possible to flag trials for further investigation where reported SAE results vary substantially from what was expected. If other safety results can be accurately modeled by learning from similar study arms in completed trials, future opportunities may include the construction of synthetic control arms\cite{thorlund2020synthetic}.

Our aim was to predict SAEs for clinical trials from their prospective trial registrations. The specific aims were to predict will the experimental arm have a higher proportion of patients experiencing SAEs than the control arm and estimate the proportion of participants who will experience SAEs in study control arms. 

\section{Methods}
\label{sec:methods}

\subsection{Dataset and preprocessing}
\begin{figure*}[htbp]
    \centering
    \setlength{\abovecaptionskip}{0pt}
    \includegraphics[width=0.9\textwidth]{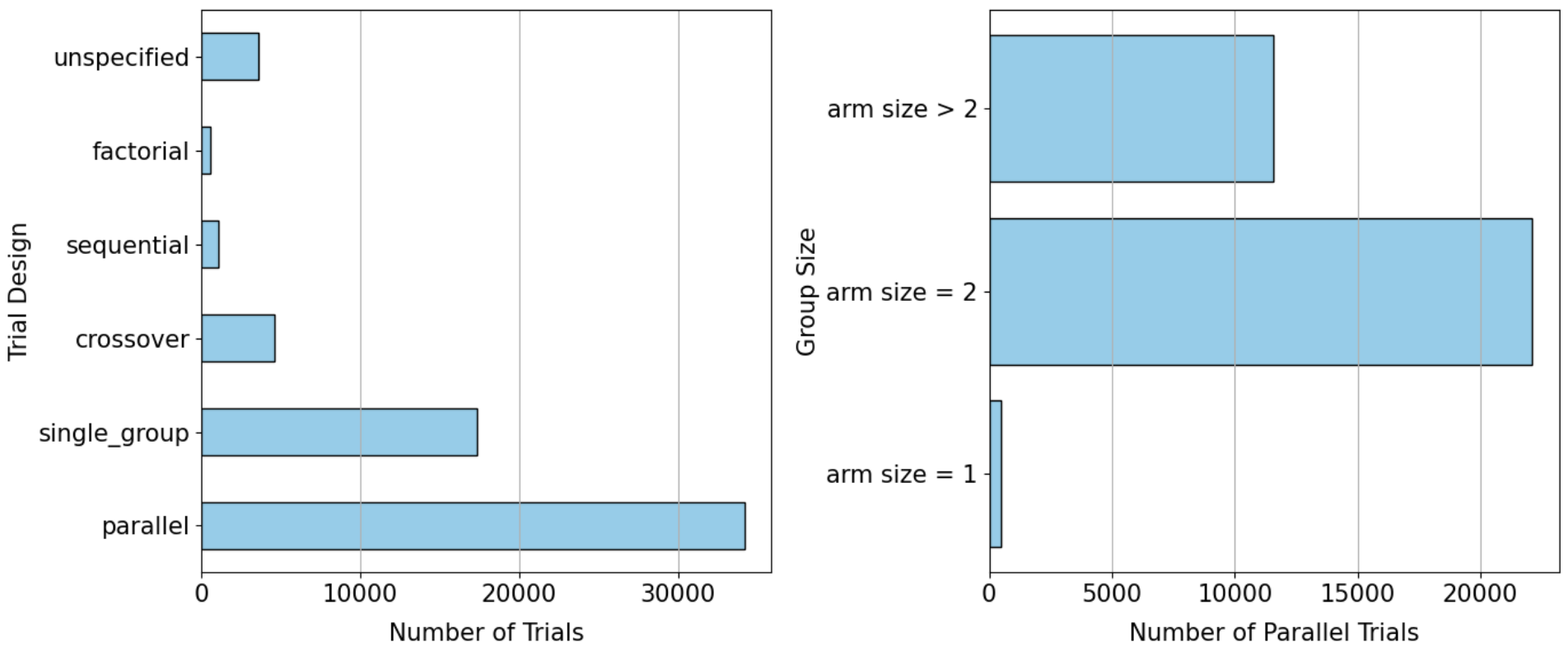}
    \caption{Distribution of 61,330 clinical trials with summary results, grouped by trial design and arm size, indicating a set of 22,107 clinical trials using a parallel group study design with two study arms as the most common type of trial.}
    \label{fig:fig1}
\end{figure*}

From a set of 292,133 interventional studies on ClinicalTrials.gov, 21\% (61,330 trials) had summary results reported on ClinicalTrials.gov (as of March 5, 2024). We included interventional studies that were parallel group clinical trials where participants were randomized to one of two groups (called trial arms), which included a set of 22,107 trials (\textbf{Figure \ref{fig:fig1}}). Clinical trials with this design typically compare a new or experimental treatment against a previous standard of care or a placebo.

Clinical trial registrations were accessed via the ClinicalTrials.gov API and formatted into a human-readable format that is suitable for language models\cite{sui2024table, deng2022turl}. The human readable format was constructed to align with the structure of the registration as in the Researcher View of the website, mimicking the view that would typically be used in appraising a trial or evidence synthesis tasks. Numbers were converted into their text equivalents (e.g. five thousand rather than 5000), to avoid known issues associated with the use of language models\cite{spathis2024first}.

Trial arms were identified as either experimental or control using regular expressions applied to the trial title, description of the interventions, and trial arms. The approach was manually evaluated for reliability across a range of trials (see \textbf{Supplementary Material}) and then applied to the complete set of 22,107 trials. 

We also extracted factors from structured data that might be associated with differences in SAE results across trial arms. We selected factors that were used in other clinical trial outcome prediction studies\cite{elkin2021predictive, wang2022predicting, feijoo2020key}, and augmented these with other factors that might be expected to be associated with higher or lower proportions of SAEs in the treatment versus placebo study arm (\textbf{Supplementary Material Table \ref{table:rf_features}}).

For the classification task of predicting whether the experimental arm will have a higher proportion of participants experiencing SAEs compared to the control arm, we used downsampling to address data imbalance. We randomly downsample the majority class (class 0), resulting in 11,542 trials (5,771 trials in each class). For the regression task of predicting the proportion of participants experiencing a SAE in the control arm, we observed a skewed distribution with an overrepresentation of low-proportion trials. We grouped trials into ten bins based on SAE proportion and randomly sampled up to 1,000 trials per bin, yielding a final set of 5,192 trials.

\subsubsection{Model architecture}
The model architecture was based on a transfer learning approach. Following preprocessing steps for each trial, its embedding was extracted using pretrained language model (see Table \textbf{\ref{table:models}}). The language model weights were frozen during training to preserve the semantic knowledge learned from large-scale corpora and to reduce computational cost. The generated embedding was then fed into a downstream prediction model, which was used for the task-specific training by calculating loss and updating weights accordingly. The model then produced a prediction based on the processed features.

\begin{table*}[!ht]
\setlength{\abovecaptionskip}{5pt}
    \centering
    \begin{tabular}{|l|c|p{7.2cm}|} 
        \hline
        \textbf{Model Name} & \textbf{Max Context Length (tokens)} & \textbf{Description} \\ \hline
        \textbf{BioBERT} \cite{lee2020biobert} & 512 & A BERT model pre-trained on biomedical literature from PubMed abstracts and PMC full-text articles. \\ \hline
        \textbf{ClinicalBERT} \cite{alsentzer2019publicly} & 512 & A BERT model pre-trained on clinical notes from the MIMIC-III database. \\ \hline
        \textbf{ClinicalT5 Encoder} \cite{lu2022clinicalt5} & 1049 & A T5 model adapted for clinical text, offering a larger context length. We use its Encoder for embedding extraction. \\ \hline
        \textbf{BGE-m3} \cite{multi2024m3} & 8192 & A language model trained on diverse text data, representing models not specialized in the biomedical domain. \\ \hline
    \end{tabular}
    \caption{Summary of pre-trained language models used for clinical and general text processing}
    \label{table:models}
\end{table*}

For the downstream prediction model, we evaluated three different modal architecture for both classification and regression tasks: K-Nearest Neighbors (KNN), Multi-layer Perceptron (MLP), and Transformer Encoder combined with MLP. In the KNN model, predictions were based on the majority label among the k-nearest neighbors in the feature space using cosine similarity (with k = 20 for regression and k = 60 for classification, selected via grid search). The MLP model was a feedforward neural network consisting of 12 fully connected layers with non-linear activation functions. The Transformer Encoder + MLP model included a Transformer Encoder with 12 layers and 8 attention heads to capture contextual relationships within the input embedding, followed by a 3-layer MLP for final prediction. The hyperparameters (k for KNN, number of layers for MLP, and layers for Transformer Encoder + MLP) were tuned using a validation set through grid search over various configurations. 

Localized language models performance can degrade when processing lengthy input due to their context length limitation\cite{liu2023lost}. Some methods have been developed to improve how language models handle long contexts, such as changing model architecture or positional encoding strategies\cite{beltagy2020longformer, yang2019xlnet, su2024roformer, zhang2024found}. While these strategies offer improvements, they need substantial additional training or fine-tuning of the language model\cite{song2024hierarchical}. 

\begin{figure*}[!htbp]
    \centering
    \setlength{\abovecaptionskip}{0pt}
    \includegraphics[width=0.78\textwidth]{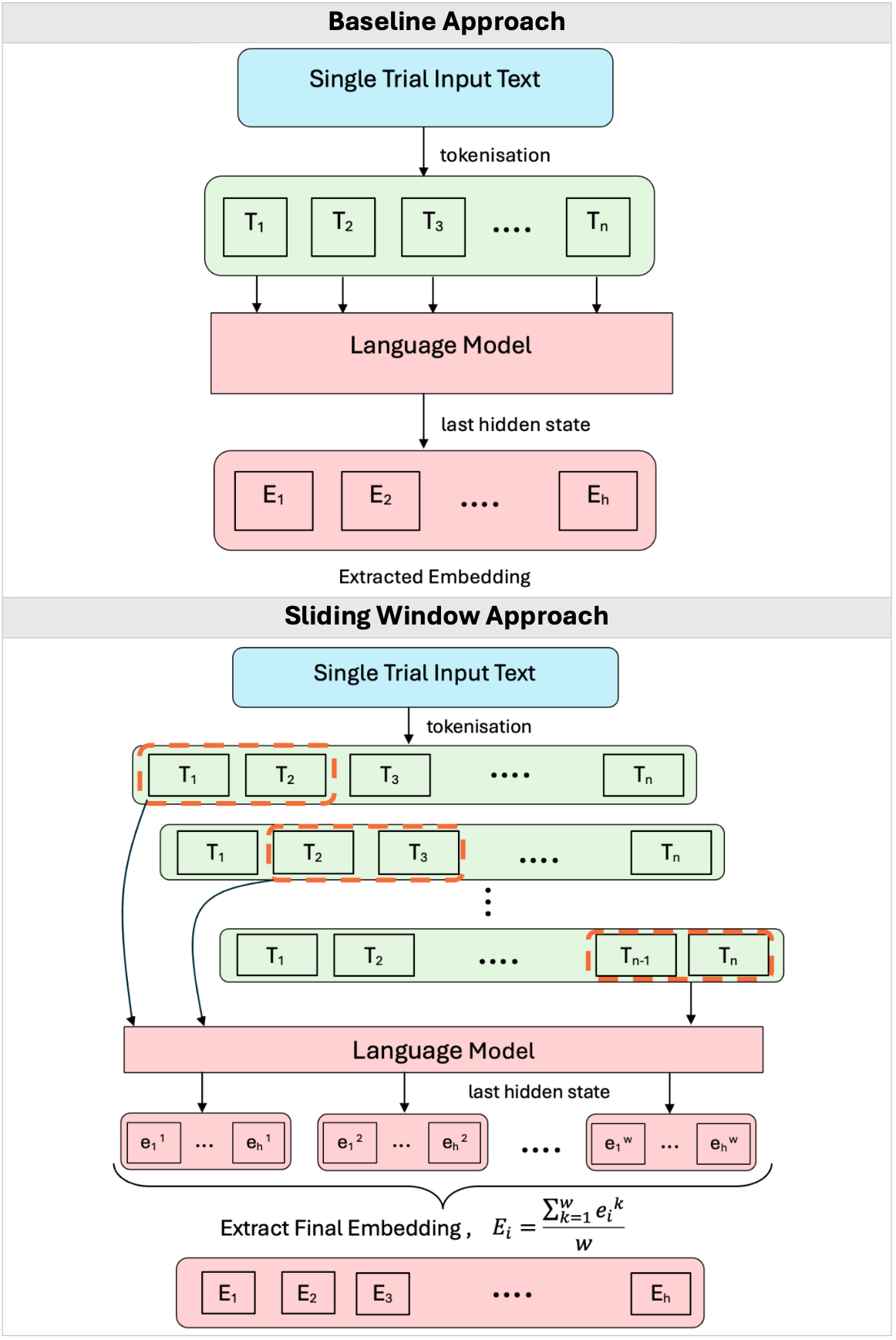}
    \caption{In a comparison example of the baseline approach and the sliding window approach, the sliding window approach a window size 2 and stride 1. The sliding window example showed above has a window size \(2\) and stride \(1\). \textcolor{blue}{\( T_i \)}: The \(i\)-th token after tokenizing the input text. \textcolor{blue}{\( n \)}: Total number of tokens generated from the input text. \textcolor{blue}{\( h \)}: Number of units in the last hidden state of the language model (e.g., 768 for BERT). \textcolor{blue}{\( w \)}: Total number of chunks generated by the sliding window. \textcolor{blue}{\( {e_{i}}^k \)}: The \(i\)-th element of the embedding from chunk \(k\). \textcolor{blue}{\( E_i \)}: The \(i\)-th element in the final embedding.}
    \label{fig:fig2}
\end{figure*}

Our feature extraction process employs a sliding window approach to ensure complete document coverage despite model context limitations. It involves splitting the input text into overlapping chunks based on the model's maximum context length, then processing each chunk individually, and combining them to produce a final embedding (\textbf{Figure \ref{fig:fig2}}). This allows us to process the long registration document while avoiding the need for extensive retraining of the language model. For instance, with BioBERT's 512-token limit, we used a 256-token stride, allowing half of the content from each chunk to carry over into the next one. We selected 50\% overlap to balance computational efficiency with context preservation, though this parameter could be optimized in future work. To establish a baseline for our experiments, we adopted a common embedding extraction approach used in previous studies involving extracting embedding on electronic health record text and clinical trial registration text\cite{wang2022predicting, gao2024transfer}. In the common embedding approach, the entire input text of a single trial is processed as one continuous sequence.

\subsubsection{Evaluation and outcome measures}
For model training, we set the learning rate to 0.0001 for all our models. To optimize the training process, we used the AdamW optimizer\cite{loshchilov2017decoupled}. We used the Cross Entropy Loss function during training for classification task and Mean Square Error Loss function for regression task training. Hyperparameters for the model architecture were chosen based on grid search through various configuration. Each classification model was trained for 20 epochs, and each regression model was trained for 40 epochs, all chosen based on the elbow point observed during training. To ensure reproducibility, we used a random seed of 42 for train and test set split, classification train set downsampling, and test set bootstrap resampling.

Accuracy, F1 Score and AUC (Area Under the Receiver Operating Characteristic Curve) were used as measures of performance in the classification task. For the regression task, performance was measured using mean absolute error (MAE; the average absolute difference between the predicted and actual values) and root mean square (RMSE; the square root of the average squared differences between the predicted and actual values), which penalizes larger errors more heavily.

The comparison of the performance for the sliding window approach is compared to the baseline approach using Wilcoxon signed-rank test. We create 30 bootstrapped test sets by resampling with replacement from the original test set. For each language model and downstream prediction model structure pair (12 combinations in total), we trained models using both the sliding window and baseline approaches, generating 30 measures for each approach per combination.

\section{Results}
\label{sec:results}
From the 11,542 trials included in the evaluation of classification task, 33\% (3,860 of 11,542) used a placebo control compared to an active comparator. Phase 3 studies (30\%, 3,466 of 11,542) and Phase 2 studies (24\%, 2,717 of 11,542) were most common. Most studies had funding from industry  (56\%, 6,475 of 11,542), most had a trial site in the United States (72\%, 8,343 of 11,542), and a minority had a Data Monitoring Committee in place (44\%, 5,090 of 11,542). Most factors were significantly associated with differences in SAE proportions across arms (\textbf{Table \ref{table:trial_sae_associations}}).

\begin{table}[!htbp]
    \centering
    \begin{tabular}{p{4.5cm} p{2.5cm} p{2.5cm} p{4.5cm}} 
        \toprule
        \textbf{Group} & \textbf{Odds Ratio (95\% CI)} & \textbf{P-value, Fisher’s Exact Test} & \textbf{\centering Trials had higher proportion of SAE from the Group (\%)} \\
        \midrule
        Trial with Placebo & 1.44 (1.33, 1.55) & \textbf{$8.77 \times 10^{-20}$} & 2161 of 3860 (56\%) \\
        Trial in Phase 1 & 0.44 (0.36, 0.53) & \textbf{$2.57 \times 10^{-19}$} & 197 of 514 (38\%) \\
        Trial in Phase 2 & 1.08 (0.99, 1.19) & $9.94 \times 10^{-2}$ & 1597 of 2717 (59\%) \\
        Trial in Phase 3 & 1.78 (1.63, 1.95) & \textbf{$3.53 \times 10^{-37}$} & 2276 of 3466 (66\%) \\
        Trial in Phase 4 & 0.53 (0.48, 0.59) & \textbf{$1.59 \times 10^{-32}$} & 799 of 1775 (45\%) \\
        Trial related to Cancer & 2.70 (2.43, 2.99) & \textbf{$7.05 \times 10^{-85}$} & 1404 of 2019 (70\%) \\
        Trial hiring Female-only patients & 0.76 (0.67, 0.86) & \textbf{$1.79 \times 10^{-5}$} & 472 of 1079 (44\%) \\
        Trial hiring Male-only patients & 0.98 (0.79, 1.21) & $8.72 \times 10^{-1}$ & 177 of 358 (49\%) \\
        Trial hiring Healthy volunteers & 0.35 (0.32, 0.39) & \textbf{$9.43 \times 10^{-85}$} & 530 of 1814 (29\%) \\
        Trial has Data Monitoring Committee & 1.67 (1.54, 1.80) & \textbf{$6.33 \times 10^{-38}$} & 2825 of 5090 (56\%) \\
        Small trial ($<$100) & 0.42 (0.39, 0.46) & \textbf{$1.03 \times 10^{-113}$} & 2048 of 5307 (39\%) \\
        Large trial ($>$100) & 2.36 (2.19, 2.55) & \textbf{$1.03 \times 10^{-113}$} & 3687 of 6172 (60\%) \\
        Trial with Industry Funding & 3.04 (2.81, 3.28) & \textbf{$6.98 \times 10^{-185}$} & 4006 of 6475 (62\%) \\
        Trial Location in USA & 0.94 (0.87, 1.02) & $1.40 \times 10^{-1}$ & 4138 of 8348 (50\%) \\
        Trial Masking (Double Blind) & 0.98 (0.89, 1.08) & $6.82 \times 10^{-1}$ & 1046 of 2111 (50\%) \\
        FDA regulated & 1.15 (1.05, 1.26) & \textbf{$2.29 \times 10^{-3}$} & 1320 of 2504 (53\%) \\
        Trial population Child Only & 0.73 (0.64, 0.85) & \textbf{$3.29 \times 10^{-5}$} & 342 of 798 (43\%) \\
        Trial population Older Adult Only & 1.15 (0.84, 1.56) & $4.30 \times 10^{-1}$ & 87 of 163 (53\%) \\
        \bottomrule
    \end{tabular}
    \caption{Structured clinical trial information and associations with higher rates of serious adverse events in the treatment/experimental arm}
    \label{table:trial_sae_associations}
\end{table}

For the classification task, a baseline random forest model using structured information showed reasonable performance (AUC: 69.27\%, F1-score: 64.25\%, Accuracy: 63.27\%), predicting whether the treatment arm had a higher proportion of patients with SAEs.

The best performing configuration for the classification task was using ClinicalT5 with the sliding window approach for embedding extraction, combined with a Transformer Encoder and MLP as the downstream prediction model structure (\textbf{Table \ref{table:classification_everything}}). The AUC was 77.58\%, F1-score 73.17\%, and accuracy 71.89\%. Statistical testing using bootstrapped Wilcoxon signed-rank tests was conducted across all three metrics, and all tests indicated significant differences in performance between configurations (\textbf{Supplementary Table \ref{table:p_values}}). 

\begin{table*}[!ht]
\setlength{\abovecaptionskip}{5pt}
    \centering
    \begin{tabular}{p{6cm} cc|cc|cc}
        \toprule
        Model Combination & \multicolumn{2}{c|}{Accuracy (\%)} & \multicolumn{2}{c|}{F1 Score (\%)} & \multicolumn{2}{c}{AUC (\%)} \\
        \cline{2-7}
                                    & Baseline & Sliding & Baseline & Sliding & Baseline & Sliding \\
        \midrule
        BioBERT + KNN               & 66.22 & 70.25 & 67.80 & 72.04 & 72.17 & 76.20 \\
        BioBERT + MLP               & 67.85 & 70.72 & 66.88 & 73.05 & 74.66 & 76.71 \\
        BioBERT + Transformer + MLP & 67.56 & 70.51 & 72.63 & 74.02 & 74.86 & 76.77 \\
        \hline
        ClinicalBERT + KNN          & 66.74 & 67.95 & 68.52 & 69.92 & 71.75 & 73.96 \\
        ClinicalBERT + MLP          & 66.96 & 69.47 & 71.11 & 70.29 & 73.55 & 75.01 \\
        ClinicalBERT + Transformer + MLP & 67.48 & 69.60 & 69.28 & 71.53 & 73.88 & 75.74 \\
        \hline
        BGEm3 + KNN                 & 66.00 & 64.62 & 67.60 & 71.12 & 71.72 & 72.61 \\
        BGEm3 + MLP                 & 68.56 & 69.34 & 68.87 & 70.30 & 73.63 & 75.89 \\
        BGEm3 + Transformer + MLP   & 68.12 & 69.73 & 69.76 & 72.53 & 74.48 & 75.87 \\
        \hline
        ClinicalT5 + KNN            & 68.37 & 69.51 & 70.41 & 71.95 & 73.87 & 75.79 \\
        ClinicalT5 + MLP            & 68.38 & 71.42 & 72.02 & \underline{\textbf{74.06}} & 75.09 & 77.29 \\
        ClinicalT5 + Transformer + MLP & 69.64 & \underline{\textbf{71.89}} & 73.02 & 73.17 & 75.77 & \underline{\textbf{77.58}} \\
        \bottomrule
    \end{tabular}
    \caption{Comparison of performance for the classification task}
    \label{table:classification_everything}
\end{table*}

The best performing configuration for the regression task was the same as the classification task (\textbf{Table \ref{table:regression_everything}}). This configuration achieved a RMSE of 18.56\%. For completeness, MAE was also reported (14.27\%), but we primarily evaluated performance based on RMSE, given that it penalizes larger errors more heavily and is more sensitive to clinically meaningful prediction deviations. Statistical testing using bootstrapped Wilcoxon signed-rank tests comparing both metrics indicated a significant difference in performance between all configurations (\textbf{Supplementary Material Table \ref{table:regression_p_values}}).

\begin{table*}[!ht]
\setlength{\abovecaptionskip}{5pt}
    \centering
    \begin{tabular}{p{6cm} cc|cc}
        \toprule
        Model Combination & \multicolumn{2}{c|}{MAE (\%)} & \multicolumn{2}{c}{RMSE (\%)} \\
        \cmidrule(lr){2-3} \cmidrule(lr){4-5}
                                    & Baseline & Sliding & Baseline & Sliding \\
        \midrule
        BioBERT + KNN                        & 16.22 & 15.94 & 20.61 & 20.25 \\
        BioBERT + MLP                        & 16.83 & 13.89 & 20.83 & 19.04 \\
        BioBERT + Transformer + MLP          & 17.69 & 13.85 & 22.47 & 18.82 \\
        \hline
        ClinicalBERT + KNN                   & 16.21 & 16.10 & 20.74 & 20.25\\
        ClinicalBERT + MLP                   & 17.16 & 15.02 & 21.40 & 19.63 \\
        ClinicalBERT + Transformer + MLP     & 18.10 & 14.89 & 22.62 & 19.51 \\
        \hline
        BGEm3 + KNN                          & 15.18 & 16.11 & 19.70 & 20.14 \\
        BGEm3 + MLP                          & 14.66 & \underline{\textbf{13.71}} & 19.85 & 18.93 \\
        BGEm3 + Transformer + MLP            & 14.80 & 13.89 & 19.92 & 19.16 \\
        \hline
        ClinicalT5 + KNN                     & 16.24 & 15.90 & 20.83 & 20.03 \\
        ClinicalT5 + MLP                     & 15.99 & 14.53 & 20.71 & 18.99 \\
        ClinicalT5 + Transformer + MLP       & 17.71 & 14.27 & 22.53 & \underline{\textbf{18.56}} \\
        \bottomrule
    \end{tabular}
    \caption{Comparison of performance in the regression task}
    \label{table:regression_everything}
\end{table*}

\begin{figure*}[!htbp]
    \centering
    \setlength{\abovecaptionskip}{0pt}
    \includegraphics[width=1\textwidth]{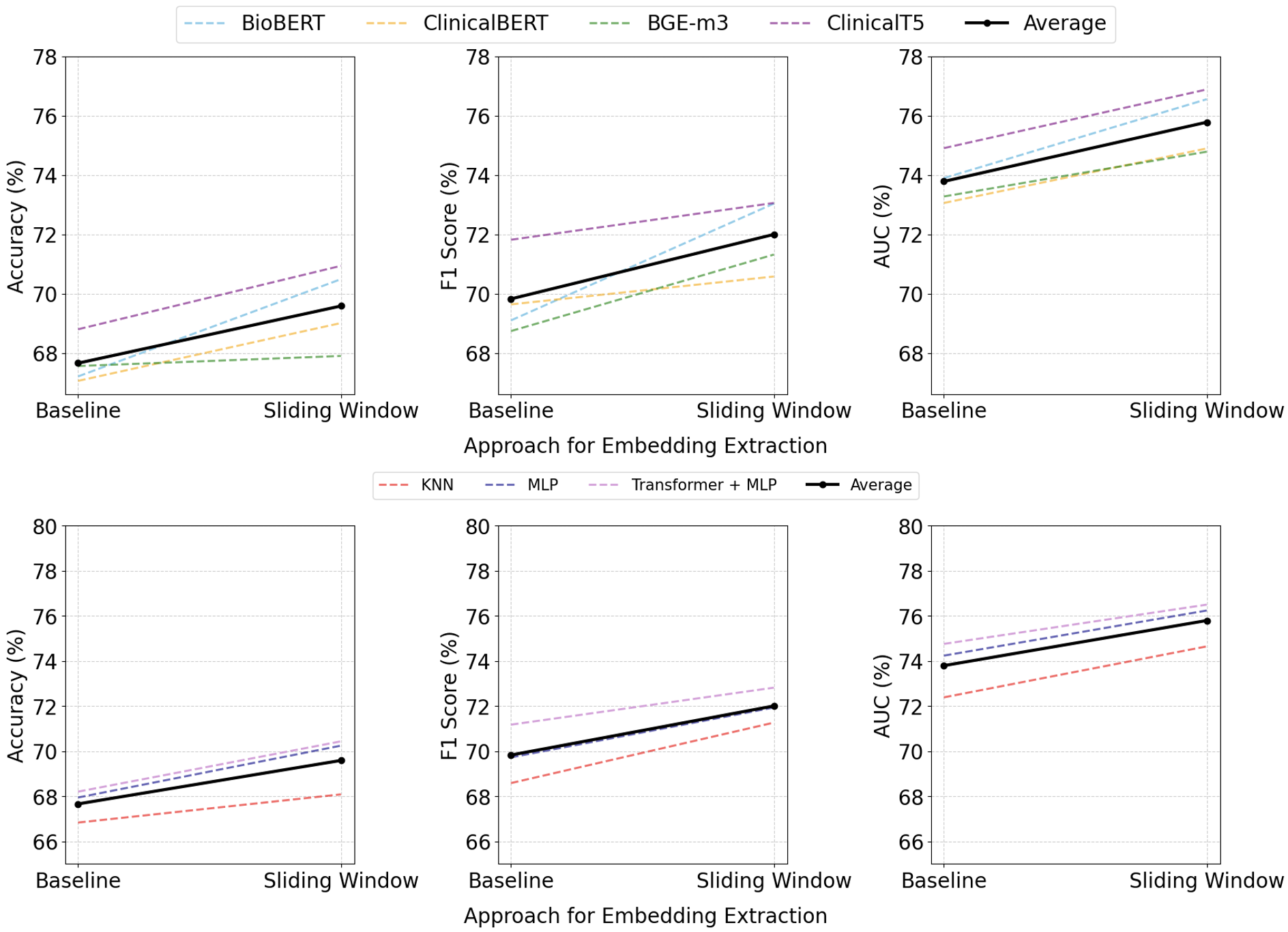}
    \caption{Performance comparison (Accuracy, F1, AUC) of different model design using baseline and sliding window approaches in the classification task. On top panel, for each language model, the metrics are averaged across the three different downstream prediction model structures. On bottom panel, for each different downstream prediction model, the metrics are averaged across four language models. The black line represents the average performance across all language models. The figure shows the consistent performance improvements achieved by using the sliding window approach compared to the baseline.}
    \label{fig:fig3}
\end{figure*}

\begin{figure*}[!htbp]
    \centering
    \setlength{\abovecaptionskip}{0pt}
    \includegraphics[width=1\textwidth]{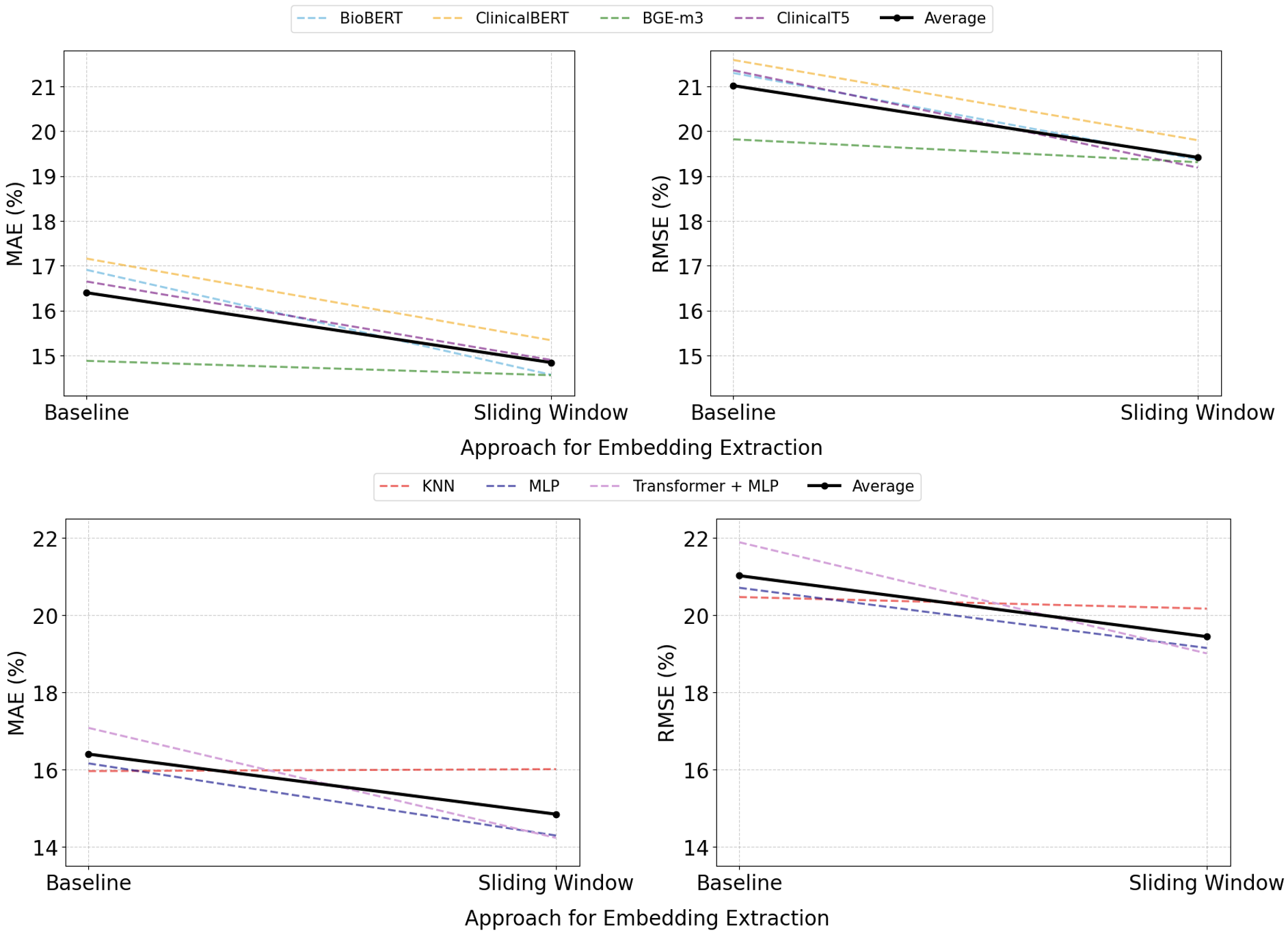}
    \caption{Performance comparison (MAE, RMSE) of different model design using baseline and sliding window approaches in the regression task. On top panel, each language model the metrics are averaged across the three different downstream prediction model structures. On bottom panel, for each different prediction model structures, the metrics are averaged across the four language models. The black line represents the average performance across all language models. The figure shows the consistent performance improvements achieved by using the sliding window approach compared to the baseline.}
    \label{fig:fig4}
\end{figure*}

Across both the classification and regression tasks, the sliding window method consistently outperformed the baseline approach. For the classification task, a comparison of the language models (averaging the performance for the three downstream prediction model structures), and comparing the downstream prediction models (averaging the performance across the language models), the results showed consistently higher performance using the sliding window approach compared to the baseline approach (\textbf{Figure \ref{fig:fig3}}). The results were comparative for the regression task, where the error measures were lower for the sliding window approach compared to the baseline approach (\textbf{Figure \ref{fig:fig4}}).

\section{Discussion}
\label{sec:discussion}
In this study we demonstrated the feasibility of predicting aspects of results of clinical trials related to SAEs, using only information from the prospective registration of the trial. The predictive models that incorporated detailed text information outperformed baseline models that only used structured information. We also found that domain pretrained models, especially those trained on biomedical or clinical texts, consistently outperformed general models in processing clinical registration data. 

Other studies have used ClinicalTrials.gov data to predict trial characteristics or outcomes. Previous work has focused on predicting trial publication likelihood (71.9\% AUC)\cite{wang2022predicting}, predicting which trials will progress to later phase trials (76.0-87.3\% AUC)\cite{feijoo2020key}, trial termination (72.6\% AUC)\cite{elkin2021predictive}, and phase 1 trial duration for specific diseases (76.8\% AUC)\cite{long2023predicting}. In contrast, our work is the first to predict SAE results, achieving 77.6\% AUC for identifying whether the experimental arm will have higher rates of SAEs compared to the control arm, and 18.6\% RMSE for predicting the proportion of participants experiencing SAEs in the control arm. These results demonstrate the feasibility of predicting safety outcomes from registration data alone. 

To handle the registration text from trials, earlier studies used traditional feature extraction methods, such as extract embeddings using Doc2Vec or TF-IDF\cite{elkin2021predictive}, or pattern matching using eligibility criteria\cite{feijoo2020key}. Another previous study took a similar approach and used pretrained language models for embedding extraction\cite{wang2022predicting}. Our approach extends this methodology in several ways. We explored variety of models across multiple architectures, pre-processed trial data into a human-readable formats, and applied a sliding window technique to handle the lengthy registration text.

The sliding window approach combined with transfer learning is relatively simple and provides a practical solution for researchers working with clinical trial registration text and may be suitable for other tasks where tasks need to make sense of longer documents. We found that the solution is particularly well-suited to clinical trial registration text, as these documents are organized into distinct sections (eligibility criteria, intervention details, outcomes, sponsor information, etc.). Unlike narrative text where understanding might require connecting information across the entire document, each section in trial registrations tends to be relatively self-contained.

The model we present here may be of value for regulatory oversight. For example, they might use predictions based on trial protocols prior to approval to identify high-risk trials that may require enhanced monitoring of reporting\cite{aspden2007preventing, klein2013postmarketing}. Another possible use case might be to compare expected results to results that are finally reported in ClinicalTrials.gov or published articles, looking for unusual discrepancies as a flag for further investigation. 

Limitations of this work come from choices made related to data processing and model performance. Our use of regular expressions to identify control and experimental arms, while practical for large-scale processing, may misclassify arms in trials with non-standard terminology, introducing noise into the training data. Additionally, the focus on two-arm parallel intervention trials does not demonstrate how our approach might work with other trial designs, such as crossover studies or sequential trials. These limitations could be addressed in future work by developing more sophisticated arm identification methods and expanding the model to handle diverse trial structures. Another limitation is the interpretability of our model predictions. Post-hoc interpretability tools such as attention maps or LIME (Local Interpretable Model-Agnostic Explanations) could be applied to address this limitation in future work\cite{vaswani2017attention, ribeiro2016should}.

The level of performance for the regression task is unlikely to be sufficient for immediate deployment in the usage scenarios described above. SAE predictions remain challenging due to their inherent complexity and because numbers can vary substantially for smaller trials, which are often where higher rates of SAEs occur. However, given that no previous work has attempted to predict safety results from registration data, this study may provide a useful baseline of a context-aware approach from which to develop new and more sophisticated approaches using domain-specific models.

During label construction for our classification task, we also applied the Chi-square test to determine statistical significance of SAE differences. We labeled trials into three classes: non-significant, or if significant, as having higher SAEs in either the control or experimental arm. However, this process resulted in a highly imbalanced dataset, with most trials classified as having non-significant differences. Statistical significance depends on multiple factors including things like the sample size which are often limited in clinical trials due to cost constraints, leading to limited statistical power for detecting differences. The imbalanced dataset limited the suitability for model fine-tuning and our proposed training approach. Future work could explore methods to address the class imbalance and develop specialized approaches for the multi-class task that enable models to better distinguish trials with statistically significant differences in SAE proportions between study arms.

\section{Conclusion}
\label{sec:conclusion}

We demonstrated that some results related to serious adverse events in clinical trials can be predicted with reasonable performance from registration information using natural language processing techniques. While the approach is currently limited to two-arm parallel trials and faces precision constraints, it shows that trial registration text contains valuable safety-related signals that can be extracted automatically. These findings suggest that as language models continue to improve, a range of new opportunities exist for estimating safety results from trial registrations, and these may provide useful information to improve the design and reporting of clinical trials.

\section{Supplementary Materials}
\setcounter{table}{0}
\subsection{Detailed information on data preprocessing}
To preprocess the raw data from ClinicalTrials.gov, the following steps were applied:
\begin{itemize}
	\item Converting JSON to Readable Format. Raw JSON data from the ClinicalTrials.gov API was converted into a human-readable report format, aligning with the structure displayed on ClinicalTrials.gov's “Researcher View”.  Language models can struggle with understanding structured data like CSV and JSON which often have limited or noisy context\cite{sui2024table, deng2022turl}. 
	\item Converting Numbers to Text. Numerical values (e.g., 5000) were converted to their textual equivalents (five thousand) to improve the language model's handling of numerical data. Embedding generated by language models for decimal numbers are often poor, and splitting numbers into separate tokens causes the loss of their numerical relationships\cite{spathis2024first, gruver2023large}.
	\item Categorizing Trial Arms. Each trial arm was labeled as either “experimental” or “control” using regular expressions based on trial title and description. This approach allowed us to efficiently identify the control and experimental arms from a large number of trials.
\end{itemize}
To validate our trial arm categorization, we manually reviewed the five trials, and each was correctly identified for control and experimental arms. 
\begin{itemize}
	\item Trial 1 (NCT01263132): The code correctly identified "F0434" as the experimental arm and "Gabapentin" as the control arm.
	\item Trial 2 (NCT01386632): The code successfully identified "DCA (dichloroacetate)" as the experimental arm and "Placebo" as the control.
	\item Trial 3 (NCT00059332): The code accurately identified "Magnesium Sulphate" as the experimental arm and "Normal Saline" as the placebo comparator.
    \item Trial 4 (NCT01904032): The code correctly identified "Vitamin D3 (50,000 IUs)" as the experimental arm and "Vitamin D3 comparator (5,000 IUs)" as the control arm.
    \item Trial 5 (NCT00004732): The code successfully identified "CAS" as the experimental arm and "CEA" as the active comparator.
\end{itemize}

\begin{table}[!h]
    \centering
    \begin{tabular}{p{3cm} p{11cm}}
        \toprule
        \textbf{Feature} & \textbf{Description} \\
        \midrule
        trial size & Total number of participants enrolled in the trial (numeric) \\ \hline
        size category & Trial size classification based on enrollment: 0=Small ($<$100 participants), 1=Large ($\geq$100 participants), -1=missing \\ \hline
        placebo & Binary indicator for whether the control arm uses placebo/sham (1) or active comparator/standard care (0) \\ \hline
        phase & Clinical trial phase encoded as ordinal values: Phase 1 (1), Phase 2 (2), Phase 3 (3), Phase 4 (4), missing (-1) \\ \hline
        cancer & Binary indicator for whether the trial studies cancer or cancer-related conditions (1=yes, 0=no) \\ \hline
        sex & Participant sex eligibility encoded as: All sexes (0), Male only (1), Female only (2), missing (-1) \\ \hline
        healthy volunteers & Binary indicator for whether the trial accepts healthy volunteers (1=yes, 0=no, -1=missing) \\ \hline
        has dmc & Binary indicator for presence of a Data Monitoring Committee (1=yes, 0=no, -1=missing) \\ \hline
        industry funded & Binary indicator for industry sponsorship or collaboration (1=industry involved, 0=non-industry only) \\ \hline
        is\_usa\_trial & Binary indicator for whether any trial site is located in the United States (1=yes, 0=no) \\ \hline
        num\_countries & Number of unique countries where the trial is conducted (numeric, 0 if missing) \\ \hline
        double\_blind & Binary indicator for double-blind masking design (1=double blind, 0=other masking types or open label) \\ \hline
        is\_fda\_regulated & Binary indicator for FDA-regulated drug or device (1=FDA regulated, 0=not FDA regulated) \\ \hline
        standard age & Standardized age categories from trial registration (0=Child only, 1=Older adult only, 2=Rest/mixed ages, -1=missing) \\ 
        \bottomrule
    \end{tabular}
    \caption{Feature used in developing the base random forest model. Missing values are encoded as -1 for all features.}
    \label{table:rf_features}
\end{table}

\subsection{Detailed P-values for classification and regression task \protect\footnote{Identical p-values arise because the Wilcoxon signed-rank test yields the same outcome whenever the signs and relative ranks of differences are consistent across comparisons. Our test has sample size 30, and the significance level is set at 0.01.}}
\begin{table}[!htbp]
    \centering
    \begin{tabular}{llcccc}
        \toprule
        \textbf{Metric} & \textbf{Model} & \textbf{bioBERT} & \textbf{clinicalBERT} & \textbf{BGEm3} & \textbf{clinicalT5} \\
        \midrule
        \multirow{3}{*}{Accuracy} 
            & MLP & \(1.86 \times 10^{-9}\) & \(3.73 \times 10^{-9}\) & \(1.04 \times 10^{-3}\) & \(1.86 \times 10^{-9}\) \\
            & Transformer + MLP & \(1.86 \times 10^{-9}\) & \(1.86 \times 10^{-9}\) & \(4.32 \times 10^{-6}\) & \(1.86 \times 10^{-9}\) \\
            & KNN (k=60) & \(1.86 \times 10^{-9}\) & \(2.76 \times 10^{-6}\) & \(4.66 \times 10^{-8}\) & \(3.90 \times 10^{-5}\) \\
        \midrule
        \multirow{3}{*}{F1} 
            & MLP & \(1.86 \times 10^{-9}\) & \(3.24 \times 10^{-6}\) & \(2.55 \times 10^{-7}\) & \(2.05 \times 10^{-7}\) \\
            & Transformer + MLP & \(1.86 \times 10^{-9}\) & \(1.86 \times 10^{-9}\) & \(1.86 \times 10^{-9}\) & \(7.61 \times 10^{-3}\) \\
            & KNN (k=60) & \(1.86 \times 10^{-9}\) & \(4.71 \times 10^{-7}\) & \(1.86 \times 10^{-9}\) & \(1.86 \times 10^{-9}\) \\
        \midrule
        \multirow{3}{*}{AUC} 
            & MLP & \(1.86 \times 10^{-9}\) & \(9.31 \times 10^{-9}\) & \(1.86 \times 10^{-9}\) & \(1.86 \times 10^{-9}\) \\
            & Transformer + MLP & \(1.86 \times 10^{-9}\) & \(1.86 \times 10^{-9}\) & \(3.73 \times 10^{-9}\) & \(1.86 \times 10^{-9}\) \\
            & KNN (k=60) & \(1.86 \times 10^{-9}\) & \(5.59 \times 10^{-9}\) & \(2.37 \times 10^{-5}\) & \(3.73 \times 10^{-9}\) \\
        \bottomrule
    \end{tabular}
    \caption{P-values for each combination of model and embedding across Accuracy, F1 Score, and AUC metrics using the Wilcoxon signed-rank test for classification between baseline and sliding window approaches}
    \label{table:p_values}
\end{table}

\begin{table}[!ht]
    \centering
    \begin{tabular}{llcccc}
        \toprule
        \textbf{Metric} & \textbf{Model} & \textbf{bioBERT} & \textbf{clinicalBERT} & \textbf{BGEm3} & \textbf{clinicalT5} \\
        \midrule
        \multirow{3}{*}{MAE} 
            & MLP & \(1.86 \times 10^{-9}\) & \(1.86 \times 10^{-9}\) & \(1.86 \times 10^{-9}\) & \(1.86 \times 10^{-9}\) \\
            & Transformer + MLP & \(1.86 \times 10^{-9}\) & \(1.86 \times 10^{-9}\) & \(1.86 \times 10^{-9}\) & \(1.86 \times 10^{-9}\) \\
            & KNN (k=20) & \(1.60 \times 10^{-5}\) & \(3.45 \times 10^{-4}\) & \(1.86 \times 10^{-9}\) & \(2.61 \times 10^{-8}\) \\
        \midrule
        \multirow{3}{*}{RMSE} 
            & MLP & \(1.86 \times 10^{-9}\) & \(1.86 \times 10^{-9}\) & \(1.86 \times 10^{-9}\) & \(1.86 \times 10^{-9}\) \\
            & Transformer + MLP & \(1.86 \times 10^{-9}\) & \(1.86 \times 10^{-9}\) & \(1.86 \times 10^{-9}\) & \(1.86 \times 10^{-9}\) \\
            & KNN (k=20) & \(3.79 \times 10^{-6}\) & \(3.73 \times 10^{-9}\) & \(4.66 \times 10^{-8}\) & \(1.86 \times 10^{-9}\) \\
        \bottomrule
    \end{tabular}
    \caption{P-values for each combination of model and embedding across MAE and RMSE metrics using the Wilcoxon signed-rank test for regression between baseline and sliding window approaches.}
    \label{table:regression_p_values}
\end{table}

\bibliographystyle{unsrtnat}
\bibliography{main}  






\end{document}